\documentclass[11pt]{article}
\pdfoutput=1
\usepackage{coling2016}
\usepackage{times}
\usepackage{url}
\usepackage{latexsym}
\usepackage{amsmath}
\usepackage{graphicx}
\usepackage{CJKutf8}
\newcommand{\seq}[1]{\mathbf{#1}}
\newcommand{\softmax}{\text{softmax}}

\title{Neural Machine Translation with Supervised Attention}

\author{Lemao Liu, Masao Utiyama, Andrew Finch and Eiichiro Sumita\\
  National Institute of Information and Communications Technology (NICT) \\
  3-5 Hikari-dai, Seika-cho, Soraku-gun, Kyoto, Japan \\
  {\tt\{lmliu,first.last\}@nict.go.jp} \\
  }
\date{}

\begin{document}
\begin{CJK}{UTF8}{gbsn}

\maketitle
\begin{abstract}
The attention mechanisim is appealing for neural machine translation,
since it is able to dynamically encode a source sentence by generating a alignment between a target word and source words.
Unfortunately, it has been proved to be worse than conventional alignment models in aligment accuracy.
In this paper, we analyze and explain this issue from the point view of reordering, and propose a supervised attention which is 
learned with guidance from conventional alignment models. Experiments on two Chinese-to-English translation tasks 
show that the supervised attention mechanism yields better alignments
leading to substantial gains over the standard attention based NMT.
\end{abstract}

\section{Introduction}
\label{intro}

Neural Machine Translation (NMT) has achieved great successes on machine translation
tasks recently \cite{bahdanau+:2014,sutskever+:2015}. Generally, it relies on a recurrent neural network under the
Encode-Decode framework: it firstly encodes a source sentence into context vectors and then generates 
its translation token-by-token, selecting from the target vocabulary.
Among different variants of NMT, attention based NMT, which is the focus of this paper, is attracting increasing interests in the community \cite{bahdanau+:2014,luong+:2015}.
One of its advantages is that it is able to dynamically make use of the encoded context through an attention mechanism 
thereby allowing the use of fewer hidden layers while still maintaining high levels of translation performance.

An attention mechanism is designed to predict the alignment of a target word with respect to source words. 
In order to facilitate incremental decoding, it tries to make this alignment prediction  
without any information about the target word itself, and thus this attention can be considered to be a form of a reordering model
(see \S \ref{rnmt} for more details).  However, it differs from conventional alignment models that are able to use the target word to infer
its alignments \cite{och+ney:2000,dyer+:2013,liu+sun:2015}, and as a result there is a substantial gap in quality between the alignments derived by this attention based NMT and conventional alignment models 
(54 VS 30 in terms of AER for Chinese-to-English as reported in \cite{cheng+:2016}). 
This discrepancy might be an indication that the potential of NMT is limited. 
In addition, the attention in NMT is learned in an unsupervised manner without explicit prior knowledge about alignment.\footnote{
We do agree that NMT is a supervised model with respect to translation rather than reordering.} 
In contrast, in conventional statistical machine translation (SMT), it is standard practice to learn reordering models in a supervised manner with the guidance 
from conventional alignment models. 

Inspired by the supervised reordering in conventional SMT, in this paper, we propose a {\em Supervised Attention} based NMT (SA-NMT) model. 
Specifically, similar to conventional SMT, we first run off-the-shelf aligners (GIZA++ \cite{och+ney:2000} or fast\_align \cite{dyer+:2013} etc.) 
to obtain the alignment of the bilingual training corpus in advance.
Then, treating this alignment result as the supervision of attention, we jointly learn attention and translation, both in supervised 
manners. Since the conventional aligners delivers higher quality alignment, it is expected that the alignment in the supervised attention NMT 
will be improved leading to better end-to-end translation performance. 
One advantage of the proposed SA-NMT is that it implements the supervision of attention as a regularization in the joint training objective (\S 3.2).
Furthermore, since the supervision of attention lies in the middle of the entire network architecture rather than the top ( as in the supervision of translation (see Figure 1(b)), 
it serves to mitigate the vanishing gradient problem during the back-propagation \cite{szegedy+:2015}. 

This paper makes the following contributions:
\begin{itemize}

\item 
It revisits the attention model from the point view of reordering (\S 2), and propose a supervised attention
for NMT that is supervised by statistical alignment models (\S 3). 
The proposed approach is simple and easy to be implemented, and it is generally applicable to any attention-based NMT models, 
although in this case it is implemented on top of the model in \cite{bahdanau+:2014}.

\item 
On two Chinese-to-English translation tasks, it empirically shows that the proposed approach gives rise to improved performance (\S 4): 
on a large scale task, it outperforms three baselines including a state-of-the-art Moses, 
and leads to improvements of up to 2.5 BLEU points over the strongest baseline; 
on a low resource task, it even obtains about 5 BLEU points over the attention based NMT system on which is it based.

\end{itemize}

\section{Revisiting Neural Machine Translation}
\label{rnmt}

\begin{figure*}[ht]
\centering
\begin{tabular}{cc}
\includegraphics[width=4.5cm]{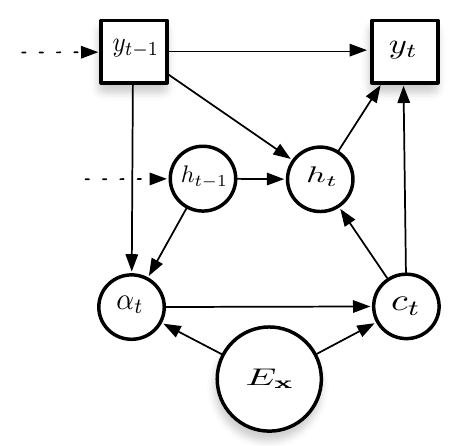} & 
\includegraphics[width=4.5cm]{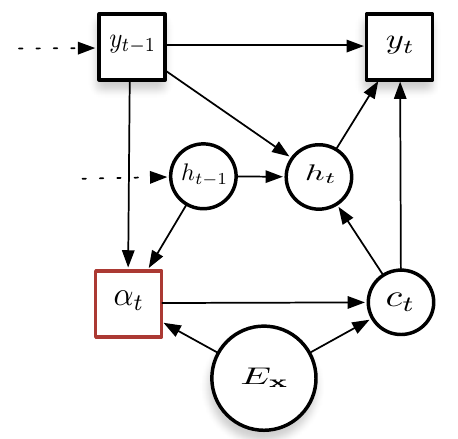} \\ 
(a) { NMT } & 
(b) { SA-NMT}
\end{tabular}
\caption{
The computational graphs of both (a) NMT and (b) SA-NMT at timestep $t$. 
Circles denote the hidden variables; while squares denote the 
observable variables, which receive supervision during training. The difference (marked in red) in (b) regarding to (a) is treating $\alpha_t$ as an observable variable instead of a hidden variable.
\label{fig:search_error}}
\end{figure*}

\noindent Suppose $\seq{x}=\left \langle x_1,x_2,\cdots,x_m \right \rangle$ denotes a source sentence, 
$\seq{y}=\left \langle y_1,y_2,\cdots,y_n \right \rangle$ a target sentence. In addition, let
$x_{<t}=\left \langle x_1,x_2,\cdots,x_{t-1} \right \rangle$ denote a prefix of $\seq{x}$.
Neural Machine Translation (NMT) directly maps a source sentence into a target under an encode-decode 
framework. In the encoding stage, it uses two bidirectional recurrent neural networks to encode $\seq{x}$ 
into a sequence of vectors $E_{\seq{x}}=\left \langle E_{x_1},E_{x_2},\cdots,E_{x_m} \right \rangle$, with
$E_{x_i}$ representing the concatenation of two vectors for $i_{th}$ source word from two directional RNNs.
In the decoding stage, it generates the target translation 
from the conditional probability over the pair 
of sequences $\seq{x}$ and $\seq{y}$ via a recurrent neural network parametrized by $\theta$ as follows:
\begin{equation}
p(\seq{y}\mid \seq{x}; \theta)  = \prod_{t=1}^{n}p(y_t\mid y_{<t}, E_{\seq{x}})
            = \prod_{t=1}^{n} \softmax\big(g(y_{t-1},h_t,c_t)\big)[y_t]
\label{eq-nmt}
\end{equation}
\noindent where 
$h_t$ and $c_t$ respectively denote an RNN hidden state (i.e. a vector) and a context vector at timestep $t$; 
$g$ is a transformation function mapping into a vector with 
dimension of the target vocabulary size; 
and $[i]$ denotes the $i_{th}$ component of a vector.\footnote{In that sense, 
$y_t$ in Eq.\eqref{eq-nmt} also denotes the index of this word in its vocabulary.} 
Furthermore, $h_t=f(h_{t-1},y_{t-1},c_t)$ is defined by an activation function, i.e. a Gated Recurrent Unit \cite{chung+:2014};
and the context vector $c_t$ is a dynamical source representation at timestep $t$, and calculated as the weighted sum of
source encodings $E_{\seq{x}}$, i.e. $c_t = \alpha_t^{\top} E_{\seq{x}}$.
Here the weight $\alpha_t$ implements an attention mechanism, and $\alpha_{t,i}$ is the alignment probability of
$y_t$ being aligned to $x_i$. 
$\alpha_t$ is derived through a feedforward neural network $a$ as follows: 
\begin{equation}
\alpha_t = a(y_{t-1},h_{t-1},E_{\seq{x}})
\label{att}
\end{equation}
\noindent where $a$ consists of two layers, the top one being a softmax layer. 
We skip the detailed definitions of $a$ together with $E_{\seq{x}}$, $f$ and $g$, and refer the readers to \cite{bahdanau+:2014} instead.\footnote{In the original paper, 
$\alpha_t$ is independent on the $y_{t-1}$ in Eq.\eqref{att}, $\alpha_t$ is independent on the $y_{t-1}$ in Eq.\eqref{att}, but this dependency was retained in our direct baseline NMT2.} 
Figure 1(a) shows one slice of computational graph for NMT definition at time step $t$.

To train NMT, the following negative log-likelyhood is minimized:
\begin{equation}
-\sum_i\log p(\seq{y}^i \mid \seq{x}^i; \theta)
\label{loss}
\end{equation}
\noindent where $\left \langle \seq{x}^i,\seq{y}^i \right \rangle$ is a bilingual sentence pair from a given training corpus, $p(\seq{y}^i \mid \seq{x}^i; \theta)$
is as defined in Eq.\eqref{eq-nmt}. 
Note that even though the training is conducted in a supervised manner with respect to translation, i.e., $\seq{y}$ are observable in Figure 1(a),
the attention is learned in a unsupervised manner, 
since $\alpha$ is hidden.

In Figure 1(a), $\alpha_t$ can not be dependent on $y_{t}$, as the target word $y_t$ is unknown at the timestep $t-1$ during the testing. 
Therefore, at timestep $t-1$, 
NMT firstly tries to calculate $\alpha_t$, through which NMT figures out those source words will be translated next, even though the next target word $y_t$ is unavailable. 
From this point of view, the attention mechanism plays a role in reordering and thus can be considered as a reordering model.
Unlike this attention model, conventional alignment models define the alignment $\alpha$ directly over
$\seq{x}$ and $\seq{y}$ as follows:
\begin{equation*}
p(\alpha\mid \seq{x},\seq{y})=\frac{\exp(F(\seq{x}, \seq{y}, \alpha))}{\sum_{a}\exp(F(\seq{x}, \seq{y},\alpha))} 
\end{equation*}
where $F$ denotes either a log-probability $\log p(\seq{y},\alpha\mid \seq{x})$ for a generative model like IBM models \cite{brown+:1993} or a feature function for discriminative models \cite{liu+sun:2015}.
In order to infer $\alpha_t$, alignment models can readily use the entire $\seq{y}$, of course including $y_t$ as well, thereby they can model the alignment between $\seq{x}$ and $\seq{y}$ more
sufficiently. As a result, the attention based NMT might not deliver satisfying alignments, as reported in \cite{cheng+:2016}, compared to conventional alignment models. 
This may be a sign that the potential of NMT is limited in end-to-end translation.

\section{Supervised Attention}
\label{spv}

In this section, we introduce supervised attention to improve the
alignment, which consequently leads to better translation performance
for NMT. Our basic idea is simple: similar to conventional SMT, it
firstly uses a conventional aligner to obtain the alignment on
the training corpus; then it employs these alignment results as 
supervision to train the NMT. During testing, decoding proceeds in exactly the same
manner as standard NMT, since there is no alignment supervision available for
unseen test sentences.

\subsection{Preprocessing Alignment Supervision}
As described in \S 2, the attention model outputs a soft alignment $\alpha$,
such that $\alpha_t$ is a normalized probability distribution.
In contrast, most aligners are typically oriented to grammar induction
for conventional SMT, and they usually output `hard' alignments, such as \cite{och+ney:2000}. 
They only indicate whether a target word is aligned to a source word or not, and
this might not correspond to a distribution for each target word. For
example, one target word may align to multiple source words, or no
source words at all.

Therefore, we apply the following heuristics to preprocess the hard
alignment: if a target word does not align to any source words, we
inherit its affiliation from the closest aligned word with preference
given to the right, following \cite{devlin+:2014}; if a target word is aligned to
multiple source words, we assume it aligns to each one evenly. In
addition, in the implementation of NMT, there are two special tokens `eol'
added to both source and target sentences. We assume they are aligned to
each other. In this way, we can obtain the final supervision of
attention, denoted as $\hat{\alpha}$.

\subsection{Jointly Supervising Translation and Attention}
We propose a soft constraint method to jointly supervise the translation
and attention as follows:

\begin{equation}
-\sum_i\log p(\seq{y}^i \mid \seq{x}^i; \theta) + \lambda \times \Delta(\alpha^i,\hat{\alpha}^i; \theta)
\label{spv-obj}
\end{equation}
\noindent where $\alpha^i$ is as defined in Eq. \eqref{eq-nmt}, $\Delta$ is a loss
function that penalizes the disagreement between $\alpha^i$ and
$\hat{\alpha}^i$, and $\lambda>0$ is a hyper-parameter that balances the
preference between likelihood and disagreement. In this way, we treat
the attention variable $\alpha$ as an observable variable as shown in Figure
1(b), and this is different from the standard NMT as shown in Figure
1(a) in essence. Note that this training objective resembles to that in multi-task learning \cite{evgeniou+pontil:2004}. 
Our supervised attention method has two further
advantages: firstly, it is able to alleviate overfitting
by means of the $\lambda$; and secondly it is capable of addressing the
vanishing gradient problem because the supervision of $\alpha$ is
more close to $E_{\seq{x}}$ than $\seq{y}$ as in Figure 1(b).

In order to quantify the disagreement between $\alpha^i$ and
$\hat{\alpha}^i$, three different methods are investigated in our
experiments:

\begin{itemize}
\item {\em Mean Squared Error} (MSE) 
\begin{equation*}
\Delta(\alpha^i,\hat{\alpha}^i; \theta) = \sum_m \sum_{n} \frac{1}{2}\big(\alpha(\theta)_{m,n}^i-\hat{\alpha}_{m,n}^i\big)^2
\end{equation*}
\noindent MSE is widely used as a loss for regression tasks \cite{lehmann+casella:1998}, and it
directly encourages $\alpha(\theta)_{m,n}^i$ to be equal to
$\hat{\alpha}_{m,n}^i$.  

\item {\em Multiplication} (MUL)
\begin{equation*}
\Delta(\alpha^i,\hat{\alpha}^i; \theta) =  -\log \big(\sum_m \sum_{n}\alpha(\theta)_{m,n}^i\times\hat{\alpha}_{m,n}^i\big)
\end{equation*}
MUL is particularly designed for agreement in word alignment and it
has been shown to be effective \cite{liang+:2006,cheng+:2016}. Note that
different from those in \cite{cheng+:2016}, $\hat{\alpha}$ is not a parametrized
variable but a constant in this paper. 

\item {\em Cross Entropy} (CE)
\begin{equation*}
\Delta(\alpha^i,\hat{\alpha}^i; \theta) = -\sum_m \sum_{n} \hat{\alpha}_{m,n}^i\times\log\alpha(\theta)_{m,n}^i 
\end{equation*}
Since for each $t$, $\alpha(\theta)_t$ is a distribution, it is
natural to use CE as the metric to evaluate the
disagreement \cite{rubinstein+kroese:2004}. 
\end{itemize}

\section{Experiments}
\label{exps}
We conducted experiments on two Chinese-to-English translation tasks: 
one is the NIST task oriented to NEWS domain, which is a large scale task and suitable to 
NMT; and the other is the speech translation oriented to travel domain, which
is a low resource task and thus is very challenging for NMT.
We used the case-insensitive BLEU4 to evaluate translation quality 
and adopted the multi-bleu.perl as its implementation.

\subsection{The Large Scale Translation Task}
\subsubsection{Preparation}
We used the data from the NIST2008 Open Machine Translation
Campaign. 
The training data consisted of 1.8M sentence
pairs, the development set was nist02 (878 sentences),
and the test sets are were nist05 (1082 sentences),
nist06 (1664 sentences) and nist08 (1357
sentences). 

We compared the proposed approach with three strong baselines:
\begin{itemize}
\vspace{-0.3cm}
\item Moses: a phrase-based machine translation system \cite{koehn+:2007};
\vspace{-0.3cm}
\item NMT1: an attention based NMT \cite{bahdanau+:2014} system at https://github.com/lisa-groundhog/GroundHog;
\vspace{-0.3cm}
\item NMT2: another implementation of \cite{bahdanau+:2014} at https://github.com/nyu-dl/dl4mt-tutorial.
\vspace{-0.3cm}
\end{itemize}
We developed the proposed approach based on NMT2, and denoted it as {\bf SA-NMT}.

\begin{table}[t]
\centering
\begin{tabular}{l|c}
Alignment Losses   & BLEU \\
\hline
\hline
Mean Squared Error (MSE) &39.4 \\
Multiplication (MUL) &39.6 \\
Cross Entropy (CE) &40.0 \\
\end{tabular}
\caption{Performance of SA-NMT on development set for different loss functions to supervise the attention in terms of BLEU.}
\label{table:losses}
\end{table}

\begin{table}[t]
\centering
\begin{tabular}{l|c}
Alignment Methods & BLEU \\
\hline
\hline
fast\_align &39.6\\
GIZA++ &40.0 \\
\end{tabular}
\caption{Comparision of aligners between fast\_align and GIZA++ for SA-NMT in terms of BLEU on the development set.}
\label{table:aligner}
\end{table}

We followed the standard pipeline to 
run Moses. GIZA++ with
grow-diag-final-and was used to build the translation
model. We trained a 5-gram target language model on
the Gigaword corpus, and used a lexicalized distortion
model. All experiments were run with the default settings.

To train NMT1, NMT2 and SA-NMT, we employed the
same settings for fair comparison. Specifically,
except the stopping iteration which
was selected using development data, we used the
default settings set out in \cite{bahdanau+:2014} for
all NMT-based systems: the dimension of word embedding
was 620, the dimension of hidden units was
1000, the batch size was 80, the source and target
side vocabulary sizes were 30000, the maximum sequence length was 50,
\footnote{This excludes all the sentences longer than 50 words in either source or target side only for NMT systems, but for Moses we use the entire training data.} 
the beam size for decoding was 12, and the optimization was 
done by Adadelta with all hyper-parameters suggested by \cite{zeiler:2012}. 
Particularly for SA-NMT, we employed a conventional word aligner to obtain the word alignment on the training data before training SA-NMT.
In this paper, we used two different aligners, which are fast\_align and GIZA++.
We tuned the hyper-parameter $\lambda$ to be 0.3 on the development set, to balance the preference between the translation and alignment.
Training was conducted on a single Tesla K40 GPU machine. 
Each update took about 3.0 seconds for both NMT2 and SA-NMT, and 2.4 seconds for NMT1. 
Roughly, it took about 10 days to NMT2 to finish 300000 updates. 

\subsubsection{Settings on External Alignments}

We implemented three different losses to supervise the attention as described in \S 3.2. 
To explore their behaviors on the development set, we employed the GIZA++ to generate the alignment on the training set prior to the training SA-NMT.
In Table \ref{table:losses}, we can see that MUL is better than MSE. 
Furthermore, CE performs best among all losses, and thus we adopt it for the following experiments.

In addition, we also run fast\_align to generate alignments as the supervision for SA-NMT and the results were reported in Table \ref{table:aligner}.
We can see that GIZA++ performs slightly better than fast\_align and thus we fix the external aligner as GIZA++ in the following 
experiments.

\subsubsection{Results on Large Scale Translation Task}

\begin{figure}[t]
\begin{center}
\includegraphics[width=7cm]{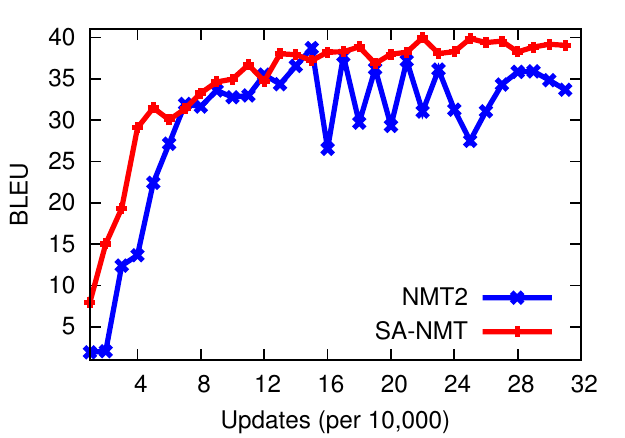}
\vspace{-0.3cm}
\caption{ Learning curves of NMT2 and SA-NMT on the development set. 
\label{fig:learn_curve} }
\vspace{-0.3cm}
\end{center}
\end{figure}

Figure \ref{fig:learn_curve} shows the learning curves of NMT2 and SA-NMT on the development set.
We can see that NMT2 generally obtains higher BLEU as the increasing of updates before peaking at
update of $150000$, while it is unstable from then on.
On the other hand, SA-NMT delivers much better BLEU for the beginning updates and 
performs more steadily along with the updates, although it takes 
more updates to reach the peaking point.

\begin{table}[t]
\centering
\begin{tabular}{c|c|ccc}
Systems  & nist02 & nist05 & nist06 & nist08\\
\hline
\hline
 Moses & 37.1 & 35.1 & 33.4 & 25.9\\
 NMT1 & 37.8 & 34.1 & 34.7 & 27.4\\
\hline
 NMT2 & 38.7  & 35.3 & 36.0 & 27.8\\
 SA-NMT & $40.0^{*}$ & $37.8^{*}$ & $37.6^{*}$ & $29.9^{*}$\\
\end{tabular}
\caption{BLEU comparison for large scale translation task. 
The development set is nist02, and the test sets are nist05,nist06 and nist08. 
`*' denotes that SA-NMT is significantly better than Moses, NMT1 and NMT2 with $p<0.01$.
Note that Moses is trained with more bilingual sentences and an additional monolingual corpus.
}
\label{table:main}
\end{table}

Table \ref{table:main} reports the main end-to-end translation results for the large scale task.
We find that both standard NMT generally outperforms Moses except NMT1 on nist05.
The proposed SA-NMT achieves significant and consistent improvements over all three baseline systems,
and it obtains the averaged gains of 2.2 BLEU points on test sets over its direct baseline NMT2. 
It is clear from these results that our supervised attention mechanism is highly effective in practice.

\subsubsection{Results and Analysis on Alignment}
\begin{figure}[t]
\begin{center}
\includegraphics[width=15.7cm]{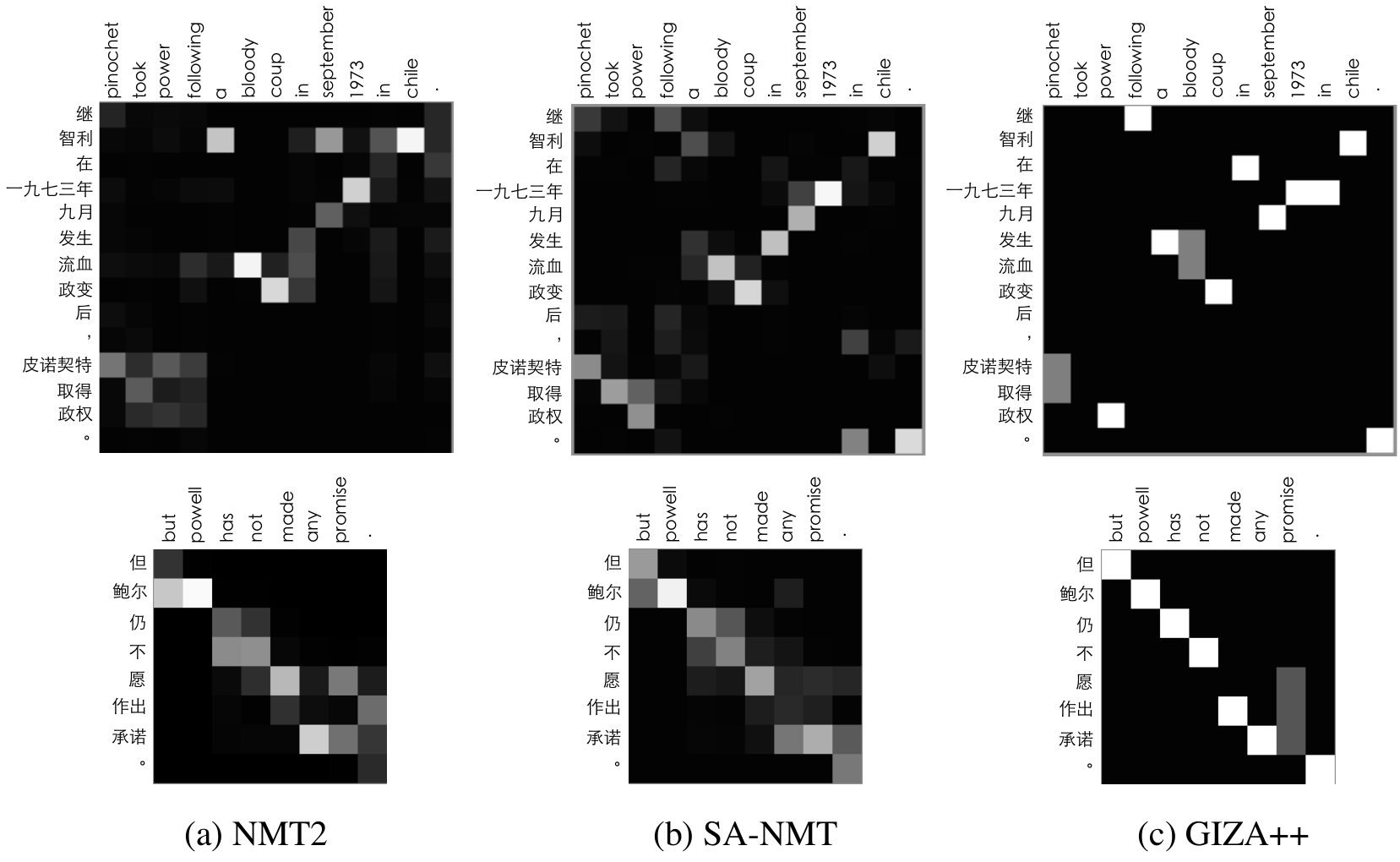} 
\caption{Example (soft) alignments of (a) NMT2 (i.e., standard NMT with unsupervised attention), (b) SA-NMT (i.e. NMT with supervised attention), 
and (c) GIZA++ on two Chinese-English sentence pairs. The soft alignments of (c) is converted from hard alignment as in \S 3.1. 
The first row shows the alignments of the sentence pair from the training set while the second row shows the alignments from test sets.
\label{fig:alignment} }
\vspace{-0.3cm}
\end{center}
\end{figure}

As explained in \S 2, standard NMT can not use the target word
information to predict its aligned source words, and thus might fail to
predict the correct source words for some target words. For example, for
the sentence in the training set in Figure \ref{fig:alignment} (a), NMT2
aligned `following' to `皮诺契特 (gloss: pinochet)' 
rather than `继 (gloss: follow)', and worse still it
aligned the word `.' to `在 (gloss: in)' rather than `。' even though this word is
relatively easy to align correctly. In contrast, with the help of information from
the target word itself, GIZA++ successfully aligned both `following' and
`.' to the expected source words (see Figure\ref{fig:alignment}(c)). With
the alignment results from GIZA++ as supervision, we can see that our
SA-NMT can imitate GIZA++ and thus align both words correctly. More
importantly, for sentences in the unseen test set, like GIZA++,
SA-NMT confidently aligned `but' and `.' to their correct source words
respectively as in Figure\ref{fig:alignment}(b), where NMT2 failed.
It seems that SA-NMT can learn its alignment behavior from
GIZA++, and subsequently apply the alignment abilities it has learned to unseen test
sentences.

\begin{table}[t]
\centering
\begin{tabular}{c|l}
Methods  & AER \\
\hline
\hline
GIZA++ & $30.6^{*}$ \\
\hline
NMT2 & 50.6  \\
SA-NMT & $43.3^{*}$ \\
\end{tabular}
\caption{Results on word alignment task for the large scale data. The evaluation metric is Alignment Error Rate (AER). 
`*' denotes that the corresponding result is significanly better than NMT2 with $p<0.01$.}
\vspace{-0.5cm}
\label{table:alignment}
\end{table}

Table \ref{table:alignment} shows the overall alignment results on word alignment task in terms of the metric, alignment 
error rate. We used the manually-aligned dataset as in \cite{liu+sun:2015} as the test set.
Following \cite{luong+manning:2015}, we force-decode both the bilingual sentences including source and reference sentences to 
obtain the alignment matrices, and then for each target word 
we extract one-to-one alignments by picking up the source word with the highest alignment confidence as the hard alignment.
From Table \ref{table:alignment}, we can see clearly that standard NMT (NMT2) is far behind GIZA++ in alignment quality.
This shows that it is possible and promising to supervise the attention with GIZA++.
With the help from GIZA++, our supervised attention based NMT (SA-NMT) significantly reduces the AER,
compared with the unsupervised counterpart (NMT2). This shows that the proposed approach is able to realize our intuition:
the alignment is improved, leading to better translation performance.

Note that there is still a gap between SA-NMT and GIZA++ as indicated in Table \ref{table:alignment}.  
Since SA-NMT was trained for machine translation instead of word alignment, 
it is possible to reduce its AER if we aim to the word alignment task only.
For example, we can enlarge $\lambda$ in Eq.\eqref{spv-obj} to bias the training objective towards word alignment task, or we can 
change the architecture slightly to add the target information crucial for alignment as in \cite{yang+:2013,tamura+:2014}.

\subsection{Results on the Low Resource Translation Task}
\begin{table}[t]
\centering
\begin{tabular}{c|c|c}
Systems  & CSTAR03 & IWSLT04 \\
\hline
\hline
Moses & 44.1 & 45.1 \\
\hline
NMT1 & 33.4  & 33.0 \\
NMT2 & 36.5  & 35.9 \\
SA-NMT & $39.8^{*}$  & $40.7^{*}$ \\
\end{tabular}
\caption{BLEU comparison for low-resource translation task.
CSTAR03 is the development set while IWSLT04 is the test set.
`*' denotes that SA-NMT is significantly better than both NMT1 and NMT2 with $p<0.01$.}
\label{table:small-main}
\end{table}

For the low resource translation task, we used the BTEC corpus as the training data, 
which consists of 30k sentence pairs with 0.27M Chinese words and 0.33M English words.
As development and test sets, we used the CSTAR03 and IWSLT04 held out sets, respectively.
We trained a 4-gram language model on the target side of training corpus for running Moses.
For training all NMT systems, we employed the same settings as those in the large scale task, 
except that vocabulary size is 6000, batch size is 16, and the hyper-parameter $\lambda=1$ for SA-NMT.

Table \ref{table:small-main} reports the final results. Firstly, we can see that 
both standard neural machine translation systems NMT1 and NMT2 are much worse than Moses with a substantial gap. 
This result is not difficult to understand: neural network systems typically require sufficient 
data to boost their performance, and thus low resource translation tasks are very challenging 
for them. Secondly, the proposed SA-NMT gains much over NMT2 similar to the case in the large scale task,
and the gap towards Moses is narrowed substantially.

While our SA-NMT does not advance the state-of-the-art Moses as in large scale translation, 
this is a strong result if we consider that previous works on low resource translation tasks: 
\newcite{arthur+:2016} gained over Moses on the Japanese-to-English BTEC corpus, 
but they resorted to a corpus consisting of 464k sentence pairs;
\newcite{luong+manning:2015} revealed the comparable performance to Moses on English-to-Vietnamese with 133k sentences pairs, 
which is more than 4 times of our corprus size. 
Our method is possible to advance Moses by using reranking as in \cite{neubig+:2015,cohn+:2016}, 
but it is beyond the scope of this paper and instead we remain it as future work.

\section{Related Work}
\label{rlw}

Many recent works have led to notable improvements in the attention mechanism
for neural machine translation. \newcite{tu+:2016} introduced an explicit coverage
vector into the attention mechanism to address the over-translation and
under-translation inherent in NMT. \newcite{feng+:2016} proposed an additional
recurrent structure for attention to capture long-term
dependencies. \newcite{cheng+:2016} proposed an agreement-based
bidirectional NMT model 
for symmetrizing alignment.
\newcite{cohn+:2016} incorporated multiple structural alignment biases
into attention learning for better alignment. All of them improved
the attention models that were learned in an unsupervised manner.
While we do not modify the attention model itself, we learn it in a
supervised manner, therefore our approach is orthogonal to theirs.

It has always been standard practice to learn reordering models from
alignments for conventional SMT either at the phrase level or word
level. At the phrase level, \newcite{koehn+:2007} proposed a lexicalized
MSD model for phrasal reordering; \newcite{xiong+:2006} proposed a
feature-rich model to learn phrase reordering for BTG; and
\newcite{li+:2014} proposed a neural network method to learn a BTG
reordering model. At the word level, \newcite{bisazza+federico:2016}
surveyed many word reordering models learned from alignment models for
SMT, and in particular there are some neural network based reordering
models, such as \cite{zhang+:2016}. Our work is inspired by these works in spirit,
and it can be considered to be a recurrent neural network based word-level reordering
model. The main difference is that in our approach the reordering model
and translation model are trained jointly rather than separately as theirs.

%
%

\section{Conclusion}
It has been shown that attention mechanism in NMT is worse than conventional word alignment models in its alignment accuracy. 
This paper firstly provides an explanation for this by viewing the atten- tion mechanism from the point view of reordering. 
Then it proposes a supervised attention for NMT with guidance from external conventional alignment models, 
inspired by the supervised reordering models in conventional SMT. 
Experiments on two Chinese-to-English translation tasks show that the proposed approach achieves better alignment results 
leading to significant gains relative to standard attention based NMT.

\section*{Acknowledgements}
We would like to thank Xugang Lu for invaluable discussions on this work.

\bibliography{coling2016}
\bibliographystyle{coling2016}
\end{CJK}
\end{document}